\documentclass[preprint,12pt]{elsarticle}

\usepackage{algorithm}
\usepackage{algorithmic}
\usepackage{newfloat}
\usepackage{listings}
\floatstyle{ruled}
\newfloat{listing}{tb}{lst}{}
\floatname{listing}{Listing}

\usepackage{amssymb}
\usepackage{amsmath}
\usepackage{graphicx}
\usepackage{booktabs}
\usepackage{xcolor}
\usepackage{multirow}
\usepackage{amsmath,amssymb,amsfonts}
\usepackage{mathrsfs}
\usepackage[title]{appendix}
\usepackage{textcomp}
\usepackage{manyfoot}
\usepackage{hyperref}
\usepackage{listings}
\usepackage{pifont}
\usepackage{soul}
\usepackage{enumitem}
\usepackage{tabularx}
\usepackage{lipsum}
\usepackage{mathbbol}
\usepackage{tabularray}
\usepackage[utf8]{inputenc}
\usepackage{color, colortbl}
\usepackage{pifont}% http://ctan.org/pkg/pifont
\newcommand{\xmark}{\ding{55}}%

\RequirePackage{algorithm}
\RequirePackage{algorithmic}
\definecolor{deepblue}{rgb}{0,0,0.5}
\newcommand*\AlgCommentInLine[1]{{\color{deepblue}{$\triangleright$ \textit{#1}}}}

\newcommand{\revise}[1]{\textcolor{black}{#1}}

\usepackage{lineno}
\journal{Image and Vision Computing}

\begin{document}

\begin{frontmatter}



\title{A2VIS: Amodal-Aware Approach to Video Instance Segmentation}

\author{
Minh Tran$^1$,
    Thang Pham$^1$,
    Winston Bounsavy$^1$,
    Tri Nguyen$^2$, Ngan Le$^1$ \\ 
    $^1$University of Arkansas, $^2$Coupang, Inc. \\
    {\small \url{https://uark-aicv.github.io/A2VIS}}
}

% \author[inst1]{Minh Tran$^\ddagger$ $^{*}$ \footnote{ $^\ddagger$ indicates corresponding author}  \footnote{$^{*}$ equal contribution}}
% \author[inst1]{Thang Pham$^{*}$}
% \author[inst1]{Winston Bounsavy}
% \author[inst2]{Tri Nguyen}
% \author[inst1]{Ngan Le }

% \affiliation[inst1]{organization={Department of Computer Science and Computer Engineering},%Department and Organization
%             addressline={1 University of Arkansas}, 
%             city={Fayetteville},
%             postcode={72701}, 
%             state={AR},
%             country={USA}}

% \affiliation[inst2]{organization={Coupang, Inc},%Department and Organization
%             city={Seattle},
%             postcode={98101}, 
%             state={WA},
%             country={USA}}

%% Abstract
\begin{abstract}
Handling occlusion remains a significant challenge for video instance-level tasks like Multiple Object Tracking (MOT) and Video Instance Segmentation (VIS). In this paper, we propose a novel framework, Amodal-Aware Video Instance Segmentation (A2VIS), which incorporates amodal representations to achieve a reliable and comprehensive understanding of both visible and occluded parts of objects in a video. The key intuition is that awareness of amodal segmentation through spatiotemporal dimension enables a stable stream of object information. In scenarios where objects are partially or completely hidden from view, amodal segmentation offers more consistency and less dramatic changes along the temporal axis compared to visible segmentation. Hence, both amodal and visible information from all clips can be integrated into one global instance prototype. To effectively address the challenge of video amodal segmentation, we introduce the spatiotemporal-prior Amodal Mask Head, which leverages visible information intra clips while extracting amodal characteristics inter clips. Through extensive experiments and ablation studies, we show that A2VIS excels in both MOT and VIS tasks in identifying and tracking object instances with a keen understanding of their full shape.
\end{abstract}


\end{frontmatter}

\section{Introduction}
\label{sec:intro}
Video Instance Segmentation (VIS) or Multiple-Object Tracking and Segmentation (MOTS) is a crucial computer vision task that entails simultaneously identifying, segmenting, and tracking all pertinent instances within a video. 
However, maintaining consistent instance tracking in VIS or MOTS encounters challenges, especially with substantial occlusions. This becomes more pronounced in long-range sequences, where instances may become heavily occluded and subsequently reappear, potentially leading to identity switches or changes \cite{heo2023generalized, zhang2023dvis}.
\begin{figure}[!t]
    \centering \includegraphics[width=.8\columnwidth]{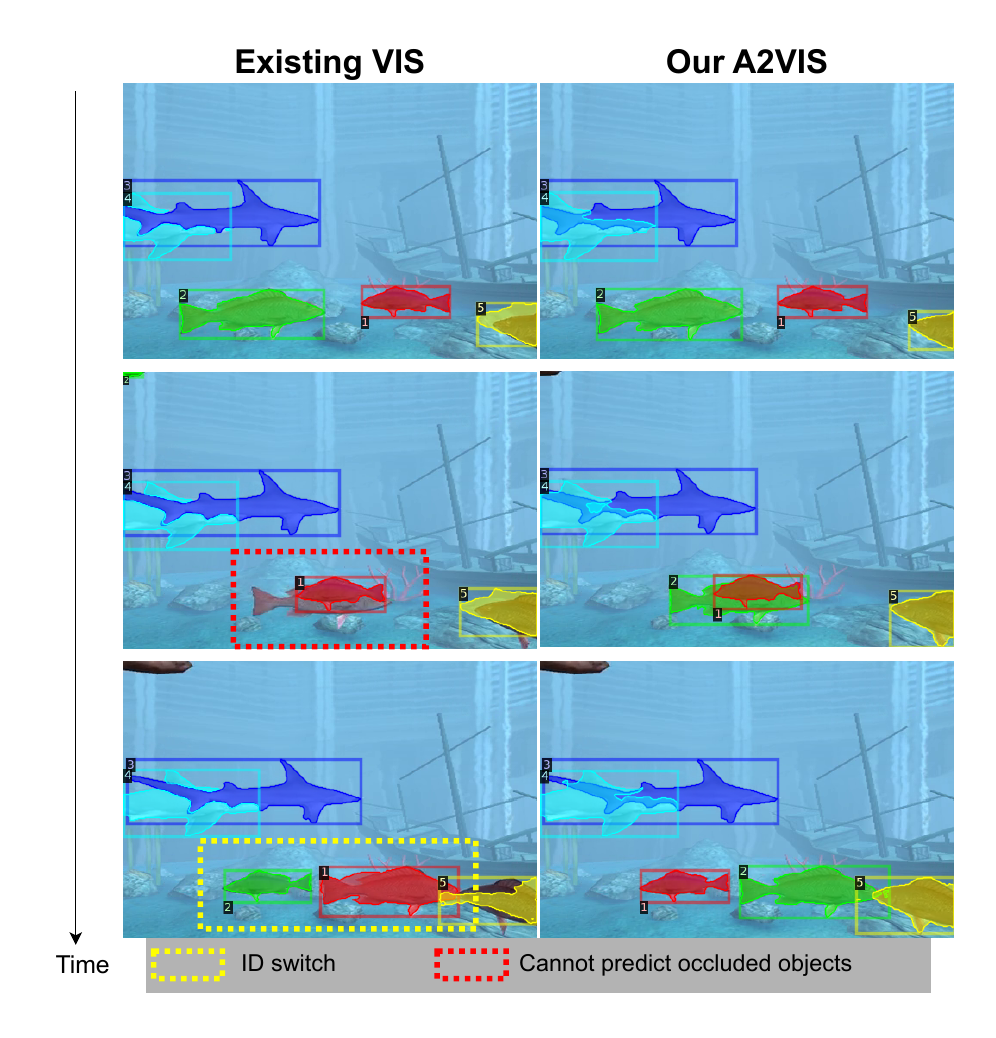}
    % \vspace{-2.3em}
    \caption{Comparison between existing VIS and the proposed A2VIS. By integrating amodal knowledge, A2VIS perceives the complete trajectory and shape of a target. This contrasts with other VIS methods that do not predict occluded parts, making them inherently susceptible to losing track of the target.}
    \vspace{-1em}
    \label{fig:teaser}
\end{figure}
Recent methods in VIS or MOT have offered various strategies to mitigate occlusion challenges such as learning inter-clip associations~\cite{heo2023generalized}, leveraging motion features~\cite{qin2023motiontrack}, \revise{\cite{zhang2023dvis}},employing tracking queries~\cite{zeng2022motr}, \revise{\cite{zhang2023dvis}}, or \revise{target-based unified approach~\cite{athar2023tarvis}}. While these methods signify progress, they primarily hinge on processing visible elements, often overlooking the comprehensive understanding of objects when parts are occluded. Higher frequencies of crossover and occlusion increase the likelihood of object identity switches. Therefore, refining the granularity of representation could be advantageous \cite{sun2022dancetrack}.

To tackle this challenge, we draw inspiration from human perception, which enables amodal perception, allowing us to perceive complete objects, even when parts of them are occluded \cite{kellman1991theory}. Recent amodal instance segmentation (AIS) \cite{li2016amodal, xiao2020amodal, tran2022aisformer, gao2023coarse} show remarkable ability on inferring complete object shapes, even in partially hidden scenarios. In light of these insights, we introduce \textbf{A}modal-\textbf{A}ware \textbf{V}ideo \textbf{I}nstance \textbf{S}egmentation (\textbf{A2VIS}), a novel framework that utilizes amodal object representation to comprehensively understand objects' shape, even when they are partially or completely hidden. Amodal segmentation experiences less dramatic changes during occlusion than visible segmentation, better maintaining object identities, as illustrated in Figure \ref{fig:teaser}. 

\revise{SAILVOS~\cite{hu2019sail} is the first paper to propose a dataset for amodal video instance segmentation. However, their proposed method on the dataset is limited to images, not videos. Specifically, their MaskJoint method is an extension of MaskRCNN~\cite{he2017mask}, featuring two mask heads—one for visible mask prediction and another for amodal prediction. While this method introduces the concept of amodal segmentation, it does not fully integrate amodal segmentation into the video instance segmentation (VIS) problem.}
Incorporating amodal representation into VIS is not straightforward, as we encounter two significant challenges: (i) effectively predicting amodal segmentation for each frame and (ii) maintaining consistent object tracking throughout the video. 
Regarding the first challenge, previous studies \cite{yao2022self, duncan1984selective} highlight the complexity of amodal segmentation, necessitating prior knowledge. Recent work attempts to model prior knowledge as shape prior \cite{xiao2020amodal, fan2023rethinking, tran2024shapeformer}. However, these methods rely on pre-training with multiple shapes of specific object types, making them dependent on type priors and difficult to generalize. To resolve that, we explore the spatiotemporal prior knowledge (SaVos~\cite{yao2022self}), which  build dense object motion across frames to explain amodal representation. 
To this end, we introduce a \textit{Spatiotemporal-prior Amodal Mask Head} (SAMH) for amodal mask prediction. Intuitively, SAMH uses two types of spatiotemporal information: short-range and long-range. Short-range information is derived from visible segments in adjacent frames. If a portion of an object is obscured in one frame, it may become visible in a neighboring frame. Long-range information involves the amodal segmentation of the object across the entire video, which is useful when an object is heavily occluded for an extended sequence of frames. To achieve this, we model these two spatiotemporal priors using a masked attention mechanism, employing a \textit{visible spatiotemporal-prior mask} (VSPM) for short-range information and an \textit{amodal spatiotemporal-prior mask} (ASPM) for long-range information. These are further elaborated in the methods section.

To address the second challenge, we introduce global instance prototypes, compressing instance representations into single embeddings to streamline detection and tracking through out the entire video. In the proposed A2VIS, these global instance prototypes capture both visible and amodal segmentation characteristics. Processed on a clip-by-clip basis, these global instance prototypes continuously associate objects from the current clip to the previous clip as well as update newly appeared objects. By encoding amodal characteristics in the global instance prototypes, the association and update procedure becomes more robust and consistent, enhancing awareness of hidden objects due to occlusion.
Our contributions can be summarized as follows:

\noindent
\textbullet{} \textbf{Novel A2VIS Framework:} We introduce A2VIS, a novel framework which utilize amodal characteristic into the processes of detection, segmentation, and tracking. A2VIS employs global instance prototypes to capture both visible and amodal characteristics of object in entire video, resulting in more robust object updates and association, especially in occluded scenarios.

\noindent
\textbullet{} \textbf{Spatiotemporal-prior Amodal Mask Head (SAMH):} We introduce a Spatiotemporal-prior Amodal Mask Head (SAMH) for predicting amodal masks by utilizing both short-range and long-range spatiotemporal information. Short-range information is derived from visible segments in nearby frames, while long-range information comes from the amodal segmentation across the entire video. These priors are modeled using a masked attention mechanism with a visible spatiotemporal-prior mask (VSPM) for short-range information and an amodal spatiotemporal-prior mask (ASPM) for long-range information.

\noindent
\textbullet{} \textbf{Performance Superiority}: Through comprehensive testing across multiple benchmarks, it is evident that A2VIS excels in identifying and tracking object instances with a keen understanding of their full shape, showing improved performance over SOTA VIS and MOT methods.

\section{Related works}

\subsection{Amodal Segmentation: }
Amodal segmentation involves predicting an object's shape, including both its visible and occluded parts, across both images and videos. While image-based amodal segmentation is usually straightforward by incorporating an occluded mask prediction ORCNN \cite{follmann2019learning}, transformer-based mask head AISFormer~\cite{tran2022aisformer}, or \revise{amodal-box expansion~\cite{liu2024blade}}, video-based amodal segmentation is more complex due to temporal consistency constraints. Approaches like SaVos~\cite{yao2022self} learn amodal representation by using visible parts from each frame and motion information via LSTM, while EoRaS~\cite{fan2023rethinking} leverages the multi-layer view fusion and temporal information to address amodal video segmentation. \revise{Recent literature has seen the emergence of diffusion models for image-based amodal segmentation. Studies such as~\cite{ozguroglu2024pix2gestalt, zhan2024amodal} leverage pretrained diffusion models for inpainting tasks to enhance the completion of occluded regions.}. \textit{Unlike existing video-based amodal segmentation approaches, which extract amodal video representation from visible masks in a multi-stage framework, A2VIS is an end-to-end framework that simultaneously detect, track, visible segmentation, and amodal segmentation for objects in videos.}

\subsection{Video Instance Segmentation (VIS): } Early VIS works like MaskTrack R-CNN~\cite{yang2019video}, SIPMask~\cite{cao2020sipmask}, SGNet~\cite{liu2021sg} extends image-based models Mask R-CNN~\cite{he2017mask} to videos by predicting frame-independent outputs and making association using post-processing during the inference stage. Later methods like IFC~\cite{hwang2021video}, Mask2Former-VIS~\cite{cheng2021mask2former}, IDOL~\cite{wu2022defense}, SeqFormer~\cite{wu2022seqformer}, MinVIS~\cite{huang2022minvis}, DVIS~\cite{zhang2023dvis} take clip-level input and run sequentially with association algorithm during post-processing. Recently, VITA~\cite{heo2022vita} introduces instance prototypes for video representation. Due to the emergence of long video benchmarks (OVIS~\cite{qi2022occluded}), those existing methods such as IFC, Mask2Former, or SeqFormer are limited in handling those benchmarks in an end-to-end manner. 
\revise{VideoCutler~\cite{oh2019video} introduces an unsupervised approach that achieves instance segmentation without relying on labeled data. OV-VIS~\cite{wang2024ov} improve open-vocabulary VIS with higher speed but maintain accuarcy. \cite{kim2024offline}  present offline-to-online knowledge dis-
tillation (OOKD) for video instance segmentation (VIS), which transfers a wealth of video knowledge from an offline model to an online model for consistent prediction. TARVIS~\cite{athar2023tarvis} presents a unified target-based segmentation approach, improves adaptability across different scenarios. }
Lately, GenVIS~\cite{heo2023generalized} addresses this by extending VITA's hypothesis with inter-clip association and criterion on instance prototypes. GenVIS utilizes a memory-based method, maintaining a single memory bank that accumulates all instance prototypes from processed clips. \textit{In contrast, A2VIS introduces a global-local instance prototype strategy. These prototypes are spatiotemporally decoded in each clip via SAMH module, enabling robust object associations, improving occlusion handling.}

\subsection{Multi-object tracking (MOT):} Addressing occlusions in MOT remains challenging. The existing works can be categorized into tracking-by-detection methods  OC-SORT~\cite{cao2023observation}, ByteTrack~\cite{zhang2022bytetrack}, FairMOT~\cite{zhang2021fairmot} and tracking-by-query-propagation methods TrackFormer~\cite{meinhardt2022trackformer}, MOTR~\cite{zeng2022motr}, MOTRv2~\cite{zhang2023motrv2},  MeMOT~\cite{cai2022memot}. The first approach first predicts the object bounding boxes for each frame, then used a separate algorithm to associate the instance bounding boxes across adjacent frames.
The second approach propose learnable queries to represent objects through out the video. The methods force each query to recall the same instance across different frames. A2VIS belong to the second category where the proposed global instance prototypes represent instances throughout the video. 
\textit{In contrast to MOT approaches using bounding boxes, which can lead to ambiguities when tracks overlap, A2VIS employs amodal segmentation. Amodal segmentation is more likely to be distinct among instances, minimizing overlapping ambiguities. Moreover, it enables the perception of entire instances even through occlusion, allowing for consistent object tracking.}

\section{Methodology}
\subsection{Problem Definition}  
In the context of traditional VIS or MOTS, we are presented with an input video denoted as \(\mathcal{V} \), comprising \(N_f\) image frames size of \( 3 \times H\times W\). These frames collectively contain \(N\) object instances observed over the duration of the video. Each instance \(i \in \{1,..,N\} \) is associated with a corresponding set of visible segmentations ${\mathbf{M}}_i$ across the frames, where ${\mathbf{M}}_i \in \mathbb{R}^{N_f \times H \times W}$. If the object \(i\) is not visibly presented in frame \(t\), \(\mathbf{{v}}_{i}[t] = \varnothing\). Otherwise, $\mathbf{{M}}_{i}[t] \in \mathbb{R}^{H \times W}$ contains the mask of instance $i$. Each instance $i$ also has a specific category $c_i$ over $C$ category predefined specifically for a dataset.

In this study, we introduce the concept of amodal segmentation to the VIS or MOTS framework. In addition to the visible segmentation, each instance \(i\) is now equipped with an additional set of amodal segmentations $\mathbf{A}_i$ across frames, where ${\mathbf{A}}_i \in \mathbb{R}^{N_f \times H \times W}$. 
Two essential considerations define the nature of amodal masks within this framework. Firstly, the amodal mask is confined within the frame size, implying that if an instance extends beyond the frame, the amodal mask will not encompass the missing parts. Secondly, \(\mathbf{A}_{i}[t'] = \varnothing\) if \(\sum_{t=1}^{t'} \mathbf{M}_i[t] = \varnothing\), meaning that if an instance has not been visibly present in the video from the start, there is no amodal mask segmentation until that time.

\subsection{Overall A2VIS}

\begin{figure*}[!t]
    \centering \includegraphics[width=\textwidth]{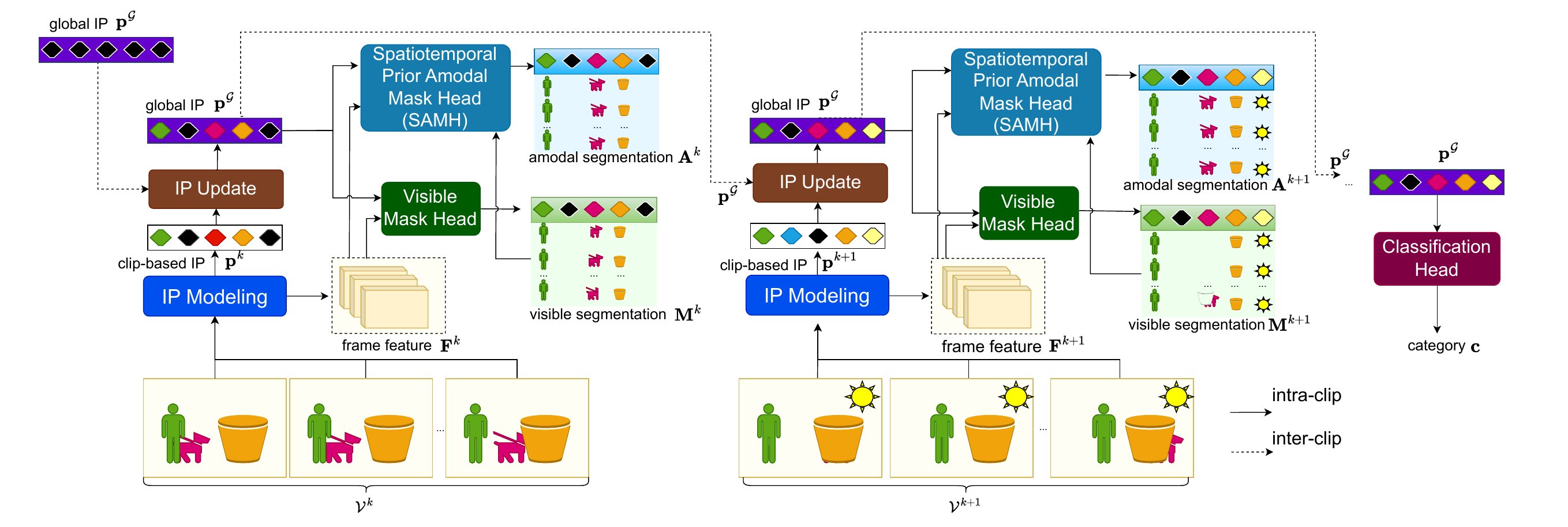}
    \vspace{-2.5em}
    \caption{Overall architecture of the proposed A2VIS. ``IP'' denotes instance prototypes in this figure. In each clip $\mathcal{V}^k$, the IP Modelling generates the clip-based IP $\mathbf{p}^k$, which is subsequently updated with the global IP $\mathbf{p}^{\mathcal{G}}$ through the IP Update module. The updated $\mathbf{p}^{\mathcal{G}}$ is then used to produce both visible segmentation $\mathbf{M}^k$ and amodal segmentation $\mathbf{A}^k$.}
    \vspace{-1em}
    \label{fig:overall_a2vis}
\end{figure*}
In this framework, we begin with an input video $\mathcal{V}$. Then we divide it into multiple clips with $N_c$ frames, $\mathcal{V}^k$, where $k \in \{1,2,.., K\}$, $K$ is the number of video clips, and $N_c < N_f$.
We define a set of global instance prototypes across clips, $\mathbf{p}^\mathcal{G} \in \mathbb{R}^{N_p \times C_e}$ that represents unique objects in the whole video $\mathcal{V}$. We have $N_p$ is the number of instance prototypes and $C_e$ represents the embedding dimension.

Initially, each clip $\mathcal{V}^k$ undergoes processing through an \textit{Instance Prototype Modelling} module to generate clip-based instance prototypes $\mathbf{p}^k$, and frame features $\mathbf{F}^k$.
Subsequently, the instance prototypes $\mathbf{p}^k $ traverse through the \textit{Instance Prototype Update} process to update the global instance prototypes $\mathbf{p}^\mathcal{G}$. This module ensures $\mathbf{p}^\mathcal{G}$ to capture new instances appearing as well as associate the local instance to the global one.
Next, the global instance prototypes $\mathbf{p}^{\mathcal{G}}$ traverses through the 
\textit{Visible Mask Head}, responsible for generating visible mask embeddings and visible segmentations specific to the video clip $\mathcal{V}^k$. 
Following this, the visible segmentation $\mathcal{V}^k$, along with the global instance prototypes $\mathbf{p}^{\mathcal{G}}$ and frame features $\mathbf{F}^k$, are processed through the \textit{Spatiotemporal Prior Amodal Mask Head} (SAMH). 
SAMH is responsible for decoding amodal characteristics for $\mathbf{p}^{\mathcal{G}}$ and predicting the corresponding amodal segmentation $\mathbf{A}^k$. 
Essentially, this module leverages the visibility of all object parts within the video clip $\mathcal{V}^k$ while also tapping into the amodal segmentation information provided by the global instance prototypes $\mathbf{p}^{\mathcal{G}}$, which accumulate amodal segmentation knowledge from the beginning. Lastly, after processing the whole video, $\mathbf{p}^{\mathcal{G}}$ is passed through the \textit{Classification Head} for predicting the instance category.
The overall of A2VIS is in Figure \ref{fig:overall_a2vis}.

\subsection{Instance Prototype Modelling}
\label{sec:IPModelling}

We adopt the object token association-based \textit{VITA} as the clip-based instance prototypes modelling $\Theta$ for its proven effectiveness and efficiency in modeling instance prototypes. This approach parses an input clip through object tokens without relying on a dense spatio-temporal backbone. It is advantageous for training on extended video sequences and facilitates establishing relationships between detected objects within the clip. Given a video clip $\mathcal{V}^k$, the model $\Theta$ returns clip-based instance prototypes $\mathbf{p}^k \in \mathbb{R}^{N_p \times C_e}$ and frame features $\mathbf{F}^k \in \mathbb{R}^{N_c\times C_e\times H_e\times W_e}$, i.e. $\{\mathbf{p}^k, \mathbf{F}^k\} = \Theta(\mathcal{V}^k)$. Here, $N_p$ is the number of clip-based instance prototypes, $C_e$ represents the embedding dimension, and $H_e$ and $W_e$ denote the spatial dimensions of the frame feature. The clip-based instance prototypes $\mathbf{p}^k$ are unique representations of objects within the video clip $\mathcal{V}^k$, each corresponding to a distinct object throughout $\mathcal{V}^k$ or representing no objects  ($\varnothing$).

\subsection{Visible Mask Head}
\label{sec:visiblemask}
In a given clip $\mathcal{V}^k$, the Visible Mask Head denoted as $\Gamma$, takes the frame feature $\mathbf{F}^k$ and the global instance prototype $\mathbf{p}^\mathcal{G}$ as inputs to generate visible segmentations $\mathbf{M}^k = \Gamma(\gamma(\mathbf{p}^\mathcal{G}), \mathbf{F}^k)$, where $\gamma$ is a visible mask embedding function implemented as a Multi-Layer Perceptron (MLP), and $\mathbf{M}^k = \{\mathbf{M}^k_i\}_{i=1}^N$ contains visible segmentations for all instance across all frames. In implementation, we define $\Gamma$ as a dot product operation to correlate visible mask embedding with the frame feature $\mathbf{F}^k$.

\subsection{Spatiotemporal-Prior Amodal Mask Head (SAMH)}
\label{sec:SAMH}

\begin{figure*}[!t]
    \centering \includegraphics[width=1\linewidth]{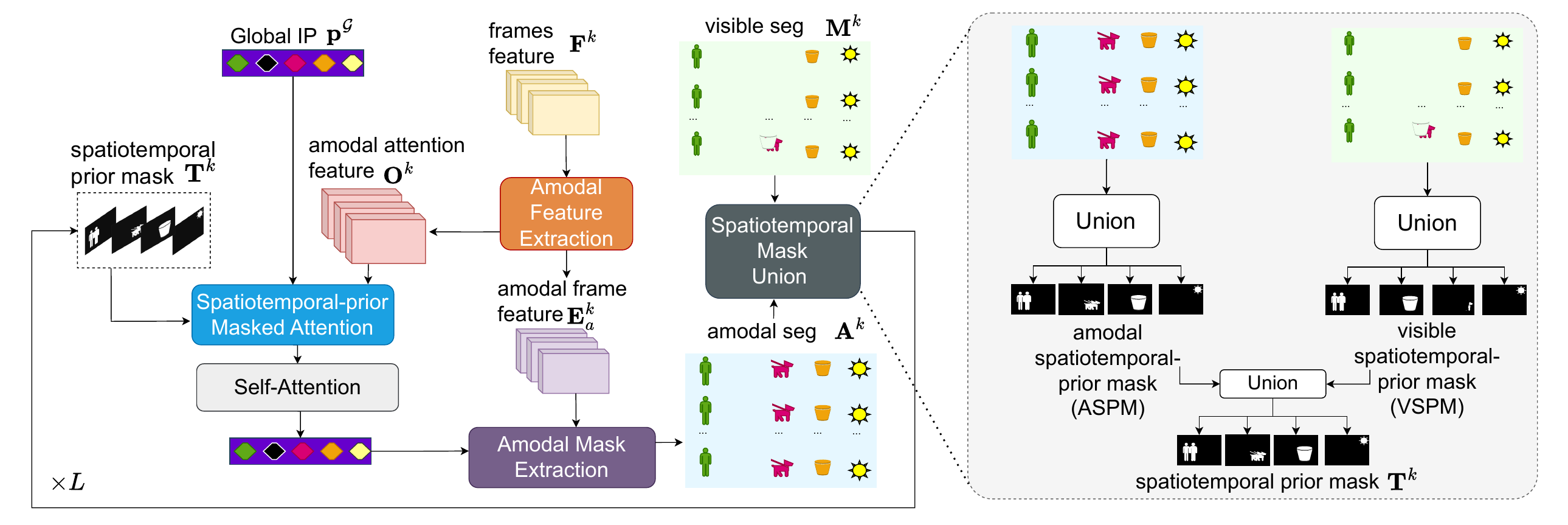}
    \vspace{-2.3em}
    \caption{Network design of Spatiotemporal-prior Amodal Mask Head (SAMH), which takes the frame feature $\mathbf{F}^k$, visible segmentation $\mathbf{M}^k$  and the global instance prototypes $\mathbf{p}^{\mathcal{G}}$ as inputs to generate amodal segmentations $\mathbf{A}^k$ and updates the global instance prototypes $\mathbf{p}^{\mathcal{G}}$. In this figure, ``IP'' denotes instance prototypes.}
    \vspace{-1em}
    \label{fig:mask_prediction}
\end{figure*}

\noindent
The objective of this module is to effectively derive amodal segmentation characteristics from each instance prototype and then predict amodal segmentation. Our approach leverages all the visible segmentation parts of an object $i$, within a video clip $\mathcal{V}^k$ as a form of visible spatiotemporal prior knowledge, play a role of short-range information. 
Additionally, we incorporate the amodal mask characteristics derived from global instance prototypes $\mathbf{p}^{\mathcal{G}}$ as the long-range information into the approach. This is particularly valuable in scenarios where the object may be occluded or not visibly present or cannot be detected within the clip.
The overall design of SAMH is illustrated in Figure \ref{fig:mask_prediction} and formally described in Algorithm \ref{algo:samh}. Within a given video clip $\mathcal{V}^k$, SAMH processes inputs that include the global instance prototypes $\mathbf{p}^{\mathcal{G}}$, frame features $\mathbf{F}^k$, and visible segmentation $\mathbf{M}^k$.

Initially, the frame features $\mathbf{F}^k$ are initially processed by an Amodal Feature Extraction $\Omega$ to obtain the amodal mask feature $\mathbf{E}^k$ and the amodal attention feature $\mathbf{O}_k$. This module can be seen as an adapter to extract the necessary amodal feature. Since amodal segmentation does not present fully in image display, the two-step generation paradigm of bootstrapping knowledge from visible mask plus prior knowledge is more effective than predicting amodal feature from scratch \cite{gao2023coarse}. Here, we follow \cite{tran2022aisformer, cheng2022masked} to design $\Omega$ by a sequence of convolutional layers ($3\times 3$ convolutional layers with a stride of $1$), where the first-half of the layers is responsible for outputting $\mathbf{O}_k$ whereas the second-half layers yields $\mathbf{E}^k$.
The amodal mask feature $\mathbf{E}^k$ serves as pixel embeddings for amodal segmentation $\mathbf{A}^k$. Meanwhile, the amodal attention feature $\mathbf{O}^k$ serves as a key-value feature, facilitating the decoding of amodal characteristics associated with the global instance prototypes $\mathbf{p}^{\mathcal{G}}$.

Following this initial processing, the decoding process proceeds with $L$ layers. 
At each layer $l \in \{1, 2, .., L\}$, the Amodal Mask Extraction function $\Phi$ takes the amodal frame feature $\mathbf{E}^k$ and the global instance prototypes $\mathbf{p}_l^\mathcal{G}$ as inputs to generate the amodal segmentation $\mathbf{A}^k = \Phi(\beta(\mathbf{p}_l^{\mathcal{G}}), \mathbf{E}^k)$. Here, $\Phi$ is defined as a dot product operation, and $\beta$ is implemented as an MLP. Subsequently, 
the  VSPM and the ASPM are computed by combining visible segmentation $\mathbf{M}^k$ and the amodal segmentation $\mathbf{A}^k$ and across $N_c$ frames within clip $\mathcal{V}^k$, respectively (Figure \ref{fig:mask_prediction} (right)). Then, 
the spatiotemporal-prior mask $\mathbf{T}^k \in \mathbb{R}^{N_p\times H_e\times W_e}$ is computed by unifying the VSPM and the ASPM.
The global instance prototypes at each layer $l$ are decoded through the proposed \textit{Spatiotemporal-prior Masked Attention} module from the previous iteration global instance prototypes $\mathbf{p}^\mathcal{G}_{l-1}$ and the amodal attention feature $\mathbf{O}^k$, given the attention mask $\mathbf{T}^k$. Formally, the Spatiotemporal-prior Masked Attention module can be expressed as follow:
\begin{subequations}
\begin{align}
    & \mathbf{p}^\mathcal{G}_l = \text{softmax} (\mathbf{T}^k + \mathbf{Q}\mathbf{K}^\top) \mathbf{V} + \mathbf{p}^\mathcal{G}_{l-1}. \\
    & \mathbf{Q} = \mathbf{p}^\mathcal{G}_l \cdot  \mathbf{W}^\mathbf{Q} \text{; } \mathbf{K} = \mathbf{O}^k  \cdot \mathbf{W}^\mathbf{K} \text{ ; } \mathbf{V}  = \mathbf{O}^k  \cdot \mathbf{W}^\mathbf{V}.
\end{align}
\end{subequations}
Here, $\mathbf{W}^\mathbf{Q}$, 
$\mathbf{W}^\mathbf{K}$,
$\mathbf{W}^\mathbf{V}$ 
are learning parameters of query $\mathbf{Q}$, key $\mathbf{K}$ and value $\mathbf{V}$, respectively. 
This attention mechanism facilitates the integration of visible prior information from adjacent frames through VSPM and incorporates amodal prior information from preceding clips via ASPM, enabling the prediction of amodal segmentation for the current frame.
Following the Spatiotemporal-prior Masked Attention, the process continues with Self-Attention, which aims to capture the correlation between instance prototypes.
After the decoding process, the final amodal segmentation $\mathbf{A}^k$ of SAMH is computed via the Amodal Mask Extraction using the final-layer decoded instance prototypes $\mathbf{p}^\mathcal{G}_{L}$.

\subsection{Instance Prototypes Update}
\label{sec:IPupdate}
While $\mathbf{p}^k$ encapsulates the instance prototypes at the clip level for a specific video clip $\mathcal{V}^k$, the global instance prototypes $\mathbf{p}^{\mathcal{G}}$ encompass all instance prototypes throughout the entire video. To update $\mathbf{p}^{\mathcal{G}}$, we utilize a traditional cross-attention mechanism as follows:
\begin{subequations}
\begin{align}
    \mathbf{Z} = (\mathbf{W}^{\mathbf{Q}'} \mathbf{p}^{\mathcal{G}})^{\top}
    \cdot \mathbf{K}'(\mathbf{p}^k). \\
    \mathbf{p}^\mathcal{G} = \mathbf{p}^\mathcal{G} + \mathbf{Z}\mathbf{W}^{\mathtt{V'}}\mathbf{p}^k.
\end{align}
\end{subequations}
Here $\mathbf{W}^{\mathbf{Q}'}, \mathbf{W}^{\mathbf{K}'}$, and $\mathbf{W}^{\mathbf{V}'}$ are learning parameters to obtain query, key, and value feature from $\mathbf{p}^{\mathcal{G}}$.

% \begin{wrapfigure}{R}{0.65\textwidth}
% \begin{minipage}{0.65\textwidth}

\begin{algorithm}[!t]
\centering
\small
\caption{Spatiotemporal-Prior Amodal Mask Head (SAMH)}
\label{algo:samh}
\begin{algorithmic}
\STATE {\bfseries Input:} {$\mathbf{F}^k$, $\mathbf{p}^\mathcal{G}$, $\mathbf{M}^k$} 
\STATE {\bfseries Output} {: $\mathbf{A}^k$, $\mathbf{p}^\mathcal{G}$}
\STATE {}
\STATE {$ \mathbf{E}^k, \mathbf{O}^k \gets \rho(\mathbf{F}^k)$} \hfill\AlgCommentInLine{Amodal Feature Extraction}
\STATE {$\mathbf{p}^\mathcal{G}_0 \gets \mathbf{p}^\mathcal{G}$} 
\FOR{$l \in \{1,2,..,L\}$}

\STATE{$\mathbf{A}^{k} \gets \Phi (\beta (\mathbf{p}^{\mathcal{G}}_{l-1}) , \mathbf{E}^k) $} \hfill\AlgCommentInLine{Amodal Mask Extraction}
\STATE{Compute the spatiotemporal-prior mask $\mathbf{T}^k$: }
\STATE{$\mathbf{T}^k \gets (\cup_{t=1}^{N_c} {\mathbf{M}}^k[t]) \cup (\cup_{t=1}^{N_c} {\mathbf{A}}^k[t])$}
\STATE{$\mathbf{T}^k(x, y) \gets \left\{\begin{array}{ll}
  0  & \text{if~} \mathbf{T}^k(x,y)=1 \\
    -\infty & \text{otherwise} \end{array}\right.$}
\STATE{$\mathbf{p}^\mathcal{G}_l \gets \operatorname{Spatiotemporal-priorMaskedAttn}(\mathbf{p}^\mathcal{G}_{l-1}, \mathbf{O}^k, \mathbf{T}^k) $}
\STATE{$\mathbf{p}^\mathcal{G}_l \gets \operatorname{SelfAttention}(\mathbf{p}^\mathcal{G}_{l})$} 
\ENDFOR
\STATE{$\mathbf{A}^k \gets \Phi (\beta (\mathbf{p}^{\mathcal{G}}_L) , \mathbf{E}^k)$}
\STATE{$\mathbf{p}^\mathcal{G} \gets \mathbf{p}^\mathcal{G}_L$}
\STATE{\textbf{return} $\mathbf{A}^k,\mathbf{p}^\mathcal{G}$}
\end{algorithmic}
\end{algorithm}

\subsection{Classification Head}
\label{sec:class_head}

At the end of the process through the whole video $\mathcal{V}$, global prototypes $\mathbf{p}^\mathcal{G}$ is passed through a Classification Head. This head is responsible for predicting the category probabilities of the instance ${\textbf{c}} \in \mathbb{R}^{N_p\times (C+1)}$, covering $C$ categories along with an auxiliary label ``no object''.  The design of this classification mask head is a class 
embedded MLP followed by a Fully-Connected (FC) layer.

\begin{table*}[!t]
\centering
    \setlength{\tabcolsep}{3pt}
    \renewcommand{\arraystretch}{1.1}
\caption{\textit{VIS tracking} comparison on FISHBOWL and SAILVOS using ResNet-50 and Swin-L backbones. \revise{For each backbone, the best results are in bold, and the second-best results are underlined.}} 
\label{tab:fishbowl_sailvos_visible}
\resizebox{\textwidth}{!}{%
    \begin{tabular}{c|l|c|cc|ccc|cc|ccc}
    \hline
        ~ & \multirow{2}{*}{\textbf{Methods}} & \multirow{2}{*}{\textbf{Backbones}} & \multicolumn{5}{c|}{\textbf{FISHBOWL}} & \multicolumn{5}{c}{\textbf{SAILVOS}} \\ \cline{4-13}
        & & & \multicolumn{2}{c|}{\textbf{Seg}} & \multicolumn{3}{c|}{\textbf{BBox}}& \multicolumn{2}{c|}{\textbf{Seg}} & \multicolumn{3}{c}{\textbf{BBox}} \\\cline{4-13}
        ~ & ~ & ~ & \textbf{AP}$\uparrow$ & \textbf{AR}$\uparrow$ & \textbf{HOTA}$\uparrow$ & \textbf{IDF1}$\uparrow$ & \textbf{IDs}$\downarrow$ & \textbf{AP}$\uparrow$ & \textbf{AR}$\uparrow$ & \textbf{HOTA}$\uparrow$ & \textbf{IDF1}$\uparrow$ & \textbf{IDs}$\downarrow$ \\  \midrule
        \multirow{4}{*}{\rotatebox{90}{Online}} & MinVIS~\cite{huang2022minvis} & ResNet-50 & 37.12 & 25.13 & 41.76 & 48.90 & 3462 & 18.63 & 15.06 & 25.43 & 22.76 & 18452 \\ 
        ~ & DVIS~\cite{zhang2023dvis} & ResNet-50 & 39.12 & 26.11 & 43.07 & 49.15 & 3811 & 20.34 & 15.28 & 28.02 & 23.76 & \underline{17332} \\ 
        ~ & \revise{STEMSeg~\cite{athar2020stem}} & \revise{ResNet-50} & \revise{37.36} & \revise{25.42} & \revise{41.58} & \revise{48.67} & \revise{3624} & \revise{18.89} & \revise{15.28} & \revise{25.67} & \revise{22.95} & \revise{18625} \\ 
        ~ & \revise{HEVIS~\cite{qin2023coarse}}& \revise{ResNet-50} & \revise{37.47} & \revise{25.33} & \revise{41.42} & \revise{48.85} & \revise{3691}& \revise{18.74} & \revise{15.45} & \revise{25.49} & \revise{22.79} & \revise{18814} \\ 
        ~ & \revise{TarVIS~\cite{athar2023tarvis}} & \revise{ResNet-50} & \revise{39.28} & \revise{25.89} & \revise{42.85} & \revise{48.92} & \revise{3956} & \revise{20.12} & \revise{\underline{15.47}} & \revise{27.79} & \revise{23.95} & \revise{17521} \\ 
        ~ & IDOL~\cite{wu2022defense}  & ResNet-50 & 39.93 & {26.53} & 42.91 & 49.14 & 3901 & 21.37 & 15.32 & 28.01 & 24.43 & 17232 \\ \cline{2-13}
        ~ & IDOL~\cite{wu2022defense}  & Swin-L & 41.22 & 28.47 & 48.74 & 55.23 & 3010 & 23.94 & 15.94 & 31.11 & 26.18 & \underline{16660} \\ \midrule
        \multirow{7}{*}{\rotatebox{90}{Offline/Semi-onl.}} & SeqFormer~\cite{wu2022seqformer}  & ResNet-50 & 36.81 & 25.23 & 41.09 & 48.62 & 3528 & 17.52 & 15.12 & 24.90 & 23.13 & 19121 \\ 
        ~ & Mask2Former-VIS~\cite{cheng2021mask2former}  & ResNet-50 & 36.17 & 25.16 & 39.96 & 47.97 & 3952 & 17.44 & 14.92 & 25.55 & 22.12 & 19301 \\ 
        ~ & VITA~\cite{heo2022vita}   & ResNet-50 & 38.15 & \underline{27.34} & 40.81 & 46.38 & 3820 & 18.32 & 15.09 & 26.54 & 23.58 & 18234 \\ 
        ~ & GenVIS~\cite{heo2023generalized}  & ResNet-50 & \underline{40.04} & 26.09 & \underline{44.08} & \underline{50.08} & \underline{3480} & \underline{21.89} & {15.41} & \underline{27.93} & \underline{24.78} & {18037} \\ 
        ~ & \textbf{A2VIS (Ours)} & ResNet-50 &\textbf{41.77} & \textbf{28.07} & \textbf{46.12} &\textbf{52.14} &\textbf{3392} & \textbf{23.12} & \textbf{15.87}& \textbf{30.04} &\textbf{25.94} & \textbf{17004}\\ \cline{2-13}
        ~ & GenVIS~\cite{heo2023generalized}  & Swin-L & \underline{43.96} & \underline{28.89} & \underline{49.62} & \underline{56.11} & \underline{2912} & \underline{24.12} & \underline{15.94} & \underline{32.44} & \underline{26.32} & {16789} \\ 
        ~ & \textbf{A2VIS (Ours)} & Swin-L & \textbf{45.77} & \textbf{30.08} & \textbf{50.45} & \textbf{58.48} & \textbf{2683} & \textbf{25.66} & \textbf{16.04} & \textbf{33.79} & \textbf{28.04} & \textbf{16043} \\ \bottomrule
    \end{tabular}
}
\end{table*}
\begin{table*}[!t]
\centering
\setlength{\tabcolsep}{4pt}
\renewcommand{\arraystretch}{1.1}
% \vspace{-0.5em}
\caption{\textit{Amodal VIS tracking} comparison on FISHBOWL and SAILVOS using ResNet-50 and Swin-L backbones. \revise{For each backbone, the best results are in bold, and the second-best results are underlined.}}
% \vspace{-0.5em}
\label{tab:fishbowl_sailvos_amodal}
\resizebox{\textwidth}{!}{%
    \begin{tabular}{l|c|cc|ccc|cc|ccc}
    \hline
        \multirow{2}{*}{\textbf{Methods}} & \multirow{2}{*}{\textbf{Backbones}} & \multicolumn{5}{c|}{\textbf{FISHBOWL}} & \multicolumn{5}{c}{\textbf{SAILVOS}} \\ \cline{3-12}
            & & \multicolumn{2}{c|}{\textbf{Seg}} & \multicolumn{3}{c|}{\textbf{BBox}}& \multicolumn{2}{c|}{\textbf{Seg}} & \multicolumn{3}{c}{\textbf{BBox}} \\\cline{3-12}
        ~ & ~ & \textbf{AP}$\uparrow$ & \textbf{AR}$\uparrow$ & \textbf{HOTA}$\uparrow$ & \textbf{IDF1}$\uparrow$ & \textbf{IDs}$\downarrow$ & \textbf{AP}$\uparrow$ & \textbf{AR}$\uparrow$ & \textbf{HOTA}$\uparrow$ & \textbf{IDF1}$\uparrow$ & \textbf{IDs}$\downarrow$ \\ \midrule
        Mask2Former-Amodal  & ResNet-50 & 30.36 & 23.76 & 42.36 & 50.35 & 3379 & 18.12 & 14.21 & 29.65 & 22.31 & 21229 \\ 
        VITA-Amodal  & ResNet-50 & 33.68 & 24.99 & \underline{48.01} & 54.62 & 3415 & 20.67 & 14.97 & 30.12 & 23.11 & 20986 \\ 
        GenVIS - Amodal  & ResNet-50 & \underline{35.47} & \underline{26.57} & 47.40 & 55.30 & \underline{3316} & 21.12 & 15.04 & \underline{30.38} & \underline{23.90} & 21343 \\ 
        AISFormer-TrackRCNN  & ResNet-50 & 34.83 & 26.31 & 47.35 & \underline{55.41} & 3407 & \underline{21.77} & \underline{15.43} & 27.04 & 25.76 & \underline{17965 }\\
        \textbf{A2VIS (Ours)} & ResNet-50 & \textbf{40.16 }& \textbf{27.41} & \textbf{49.04 }& \textbf{58.43} & \textbf{3275} & \textbf{23.41} & \textbf{15.04} & \textbf{32.12} & \textbf{26.41} & \textbf{16923} \\ \midrule
        GenVIS-Amodal  & Swin-L & \underline{40.66} & \underline{28.76} & \underline{49.43} & \underline{58.29} & \underline{3242} & \underline{22.12} & \underline{15.10} & \underline{33.42} & \underline{26.42} & \underline{17212} \\ 
        \textbf{A2VIS (Ours)} & Swin-L & \textbf{43.08} & \textbf{29.56} & \textbf{51.51} & \textbf{60.19} & \textbf{2547} &\textbf{26.02} & \textbf{16.12} & \textbf{34.55} & \textbf{28.12} & \textbf{15678} \\ \hline
    \end{tabular}
}
% \vspace{-0.5cm}
\end{table*}

\subsection{Loss Function}
Let $c^{gt}, {\mathbf{M}}^{gt}$, and ${\mathbf{A}}^{gt}$ present the ground truth categories, visible segmentation and amodal segmentation of instances in the video, respectively. 
Inspired by common practices \cite{cheng2022masked, heo2022vita, heo2023generalized}, at each optimization step,
we first find the bipartite matching between the two sets of \(N_p\) instance predictions and \(N\) ground truth object instances in a video. Let $\mathfrak{S}_N$ be a set of permutations of $N$ elements. The optimal assignment \(\hat{\sigma} \in \mathfrak{S}_N\) is computed with Hungarian matching algorithm as follow:
\begin{equation}
    \hat{\sigma} = \arg\min_{\sigma \in \mathfrak{S}_N} \sum_{i=1}^N \big[-\log \hat{\mathbf{c}}_{\sigma(i)} (c^{gt}_i) + \mathbb{1}_{c^{gt}_i \neq \varnothing} (\mathcal{L}_{v} + \mathcal{L}_{a} ) \big].
\end{equation}
where \(\mathcal{L}_v = \mathcal{L}_\text{m} ({\mathbf{M}}_{\sigma(i)}, {\mathbf{M}}^{gt}_i) \) and \(\mathcal{L}_a = \mathcal{L}_\text{m} ({\mathbf{A}}_{\sigma(i)}, {\mathbf{A}}_i^{gt})\). \(\mathcal{L}_\text{m}\) is a binary cross entropy mask loss. Subsequently, given the computed optimal assignment \(\hat{\sigma}\), the final loss $\mathcal{L}_\text{final}$ is computed for backpropagation is computed as:
\begin{equation}
    \mathcal{L}_\text{final} = \sum_{i=1}^N \big[-\log {\mathbf{c}}_{\sigma(i)} (c^{gt}_i) + \mathbb{1}_{c^{gt}_i \neq \varnothing} (\mathcal{L}'_{v} + \mathcal{L}'_{a} ) \big]
\end{equation}
where \(\mathcal{L}'_v = \mathcal{L}_\text{m} ({\mathbf{M}}_{\hat{\sigma}(i)}, {\mathbf{M}}_i^{gt}) \) and \(\mathcal{L}'_a = \mathcal{L}_\text{m} ({\mathbf{A}}_{\hat{\sigma}(i)}, {\mathbf{A}}_i^{gt})\).

\section{Experimental Results}
\subsection{Datasets, Metrics}
\noindent

\noindent
\textbf{Datasets.} We benchmark A2VIS on two datasets: \textit{FISHBOWL} \cite{tangemann2021unsupervised} comprises 10,000 training videos and 1,000 testing videos, each containing 128 frames, recorded from a WebGL aquarium demo. 
\textit{SAIL-VOS}~\cite{hu2019sail} is derived from the game GTA-V, including 160 training video and 41 testing video.

\noindent
\textbf{Metrics. }We use two types of metrics: (i) Segmentation (\textit{Seg}) tracking, evaluated with Average Precision (\textit{AP}) and Average Recall (\textit{AR}) metrics from MaskTrack R-CNN; (ii) Bounding Box (\textit{BBox}) tracking, measured with Higher Order Tracking Accuracy (\textit{HOTA}) \cite{luiten2021hota}, \textit{IDF1}\cite{ristani2016performance}, and ID switch (\textit{IDs})\cite{ristani2016performance}.

\subsection{Implementation Details}

We also follow common training procedure of previous VIS works by performing the following steps.
First, we initialize the models using COCO instances segmentation~\cite{lin2014microsoft} pretrained weights corresponding to backbones (ResNet-50~\cite{he2016deep} or SwinL~\cite{liu2021swin}). Subsequently, we pretrain our A2VIS with frame-level on FISHBOWL and SAILVOS datasets, supervised by visible segmentation mask ground truth. More specifically, the frame-level detector Mask2Former~\cite{cheng2022masked} model is pretrained on the frame-level FISHBOWL and SAILVOS as the frame-level detector. Finally, once the frame-level detectors are trained, our A2VIS are trained at the video-level, supervised by both visible and amodal segmentation ground truth.

\subsection{Baselines}
\label{sec:baseline}

\noindent
\textbf{VIS Baselines.}
We compare A2VIS against SOTA VIS approaches to assess its performance in simultaneously detecting, segmenting, and tracking objects. We include both online methods such as \textit{IDOL}\cite{wu2022defense}, \textit{MinVIS} \cite{huang2022minvis}, \revise{StemSeg~\cite{athar2020stem}, TarVIS~\cite{athar2023tarvis}, HEVIS~\cite{qin2023coarse}}, and \textit{DVIS}~\cite{zhang2023dvis}, and offline/semi-online methods like \textit{SeqFormer}, \textit{Mask2Former-VIS }, \textit{VITA} \cite{heo2022vita}, and \textit{GenVIS} \cite{heo2023generalized}, using both ResNet50 \cite{he2016deep} and Swin-L \cite{liu2021swin} backbone networks, on \textit{FISHBOWL} and \textit{SAIL-VOS} datasets.

\noindent
\textbf{Amodal VIS Baselines.}
We introduce \textit{Mask2Former-Amodal}, \textit{VITA-Amodal}, and \textit{GenVIS-Amodal}, which are extensions of SOTA VIS methods to incorporate amodal supervision by replacing visible segmentation supervision with amodal supervision. In our introduced \textit{Mask2Former-Amodal}, \textit{VITA-Amodal}, and \textit{GenVIS-Amodal}, we first initialize the model with COCO instances segmentation~\cite{lin2014microsoft} pretrained corresponding to backbones (ResNet-50~\cite{he2016deep} or Swin-L~\cite{liu2021swin}). 
Next, all the models are pretrained with frame-level FISHBOWL and SAILVOS datasets on FISHBOWL dataset with amodal segmentation ground truth. 
Finally, these models are trained on video-level supervised by amodal segmentation ground truth.
We also introduce \textit{AISFormer-TrackRCNN}, an enhanced version of AISFormer, integrated with MaskTrack R-CNN, equipped with a specialized SOTA amodal mask prediction head. This model serves as a track-by-amodal-segmentation baseline. 

\noindent 
\textbf{MOT Baselines.} 
In the context of MOT baselines, we employ query-based tracking methods including \textit{TrackFormer} \cite{meinhardt2022trackformer} and \textit{MOTR} \cite{zeng2022motr}, which share a similar conceptual foundation with instance prototypes-based VIS methods. We follow the same training procedure of these methods. First, the backbone utilized for these tracking baselines is ResNet-50. In line with their respective setups, we initialized their frame-level detector Deformable DETR~\cite{zhu2020deformable} with COCO object detection~\cite{lin2014microsoft} pretrained weights. Subsequently, we train the video-level setup also with amodal bounding box ground truth.

\subsection{Quantitative Performance Comparison}
\label{sec:comparison}
\subsubsection{Comparison with SOTA VIS methods.}
In A2VIS, tracking performance is determined by global instance prototypes, which represent both visible and amodal characteristic of the instances. 
Consequently, the predicted visible segmentation of instances derived from these global instance prototypes benefits from consistent object id, maintained through the model's amodal characteristics.
To validate this, we compare A2VIS with existing SOTA VIS methods, as shown in Table \ref{tab:fishbowl_sailvos_visible}. We assess both visible instance segmentation tracking by AP and AR metrics and conventional MOT based on bounding box tracking with HOTA, IDF1, and IDs metrics. Across all backbones and datasets, A2VIS achieves the highest performance with a significant performance gap with the second best method GenVIS. Notably, the differences in IDF1 and IDS metrics highlight A2VIS's ability to maintain consistency and accuracy in object tracking, particularly due to its amodal awareness.

\begin{figure*}[!h]
    \centering
    \includegraphics[width=\linewidth]{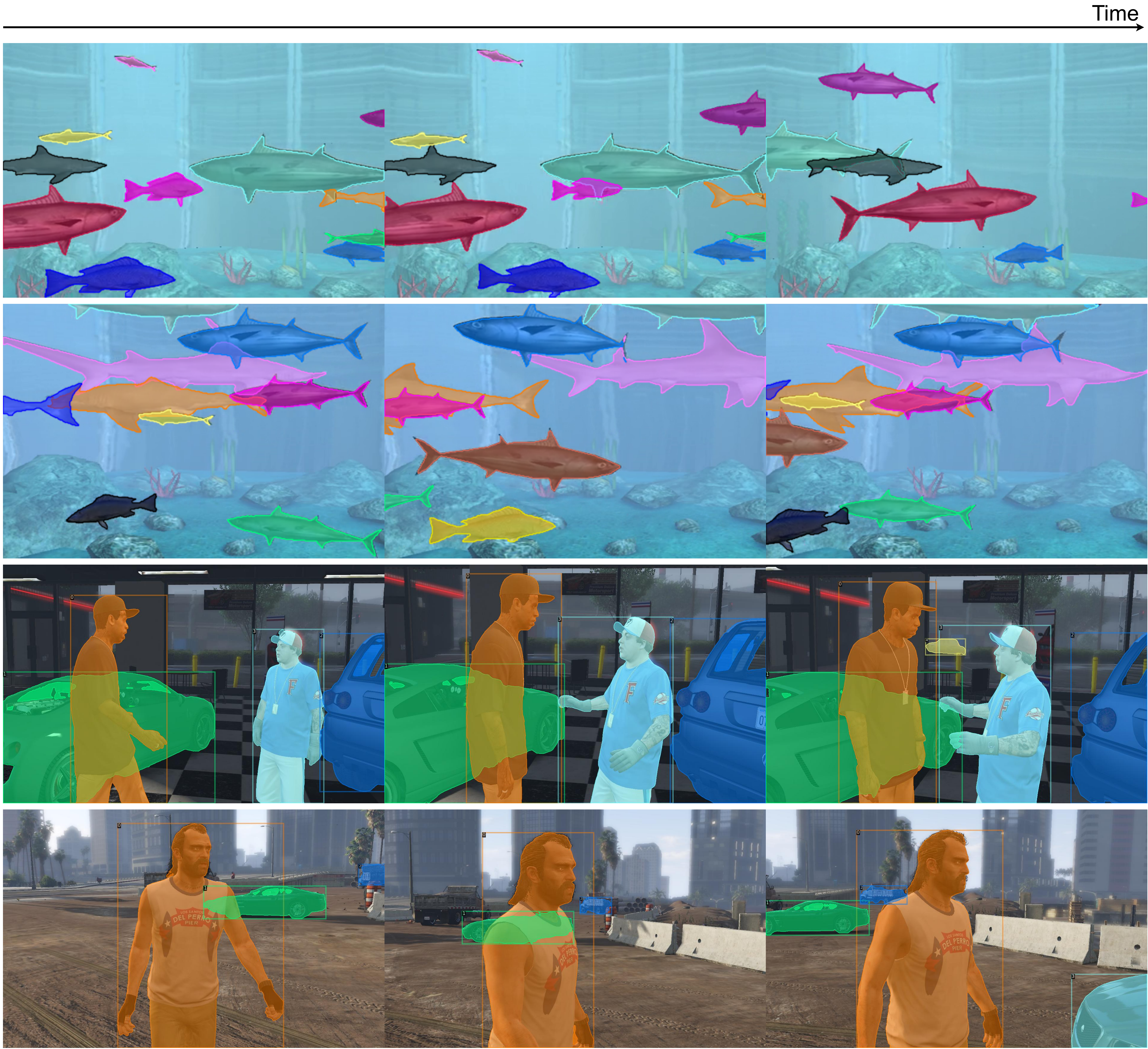}
    \caption{Qualitative results of A2VIS on FISHBOWL dataset (first two rows) and SAILVOS dataset (last two rows).}
    \label{fig:additional_quali_results}
\end{figure*}

\begin{table}[!t]
\caption{Tracking performance in comparison with MOT methods on FISHBOWL and SAILVOS using ResNet-50 backbone. All metrics are evaluated on amodal boxes. \revise{Best results are in bold, and the second-
best results are underlined.}}
% \vspace{-1em}
\setlength{\tabcolsep}{6pt}
\label{tab:track_compare}
\centering
\resizebox{\linewidth}{!}{%
\begin{tabular}{l|cccc|cccc}
    \toprule
        \multirow{2}{*}{\textbf{Method}} & \multicolumn{4}{c|}{\textbf{FISHBOWL}} & \multicolumn{4}{c}{\textbf{SAILVOS}}\\ \cline{2-9}
        ~ & \textbf{HOTA}$\uparrow$ & \textbf{DetA}$\uparrow$ & \textbf{IDF1}$\uparrow$ & \textbf{IDs} & \textbf{HOTA}$\uparrow$ & \textbf{DetA}$\uparrow$ & \textbf{IDF1}$\uparrow$ & \textbf{IDs}$\downarrow$ \\ \hline
        TrackFormer~\cite{meinhardt2022trackformer} & 42.12 & 35.03 & 54.21 & 3921 & 28.12 & 21.20 & 22.77 & 19231 \\ 
        MOTRv2~\cite{zhang2023motrv2} & \underline{47.32} & \underline{37.12} & \underline{57.45} & \underline{3391} & \underline{31.33} & \underline{25.07} & \underline{25.67} & \underline{17732} \\ \hline
        \textbf{A2VIS (Ours)} & \textbf{49.04} & \textbf{40.26} & \textbf{58.43} & \textbf{3275} & \textbf{32.12} & \textbf{24.14} & \textbf{26.41} & \textbf{16923} \\ \bottomrule
    \end{tabular}
}
% \vspace{-1.5em}
\end{table}

\begin{figure}[!t]
    \centering \includegraphics[width=.8\linewidth]{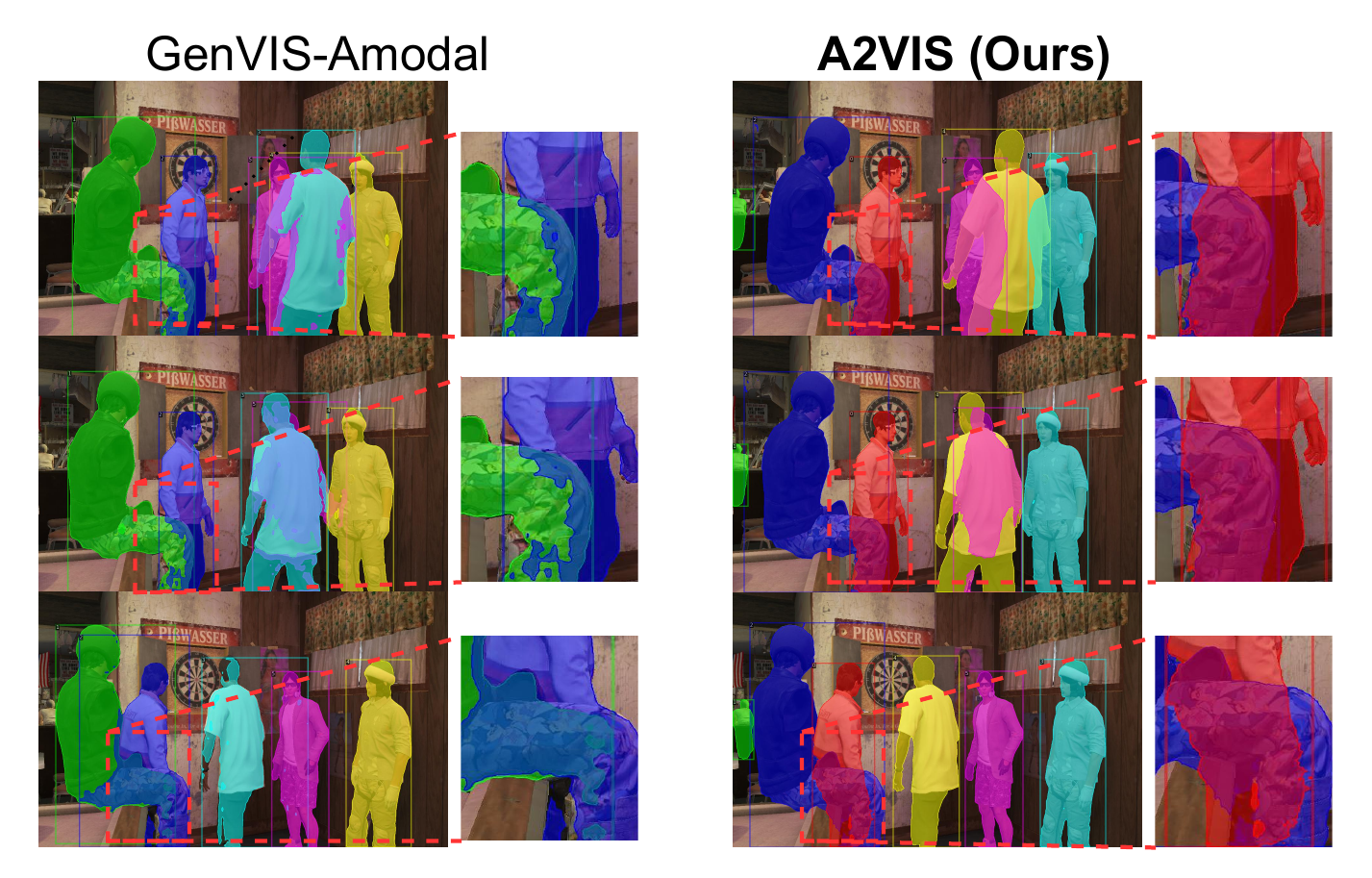}
    \caption{Qualitative comparisons of A2VIS with GenVIS-Amodal. Videos are sourced from SAIL-VOS testset.}
    \label{fig:quali_result}
\end{figure}

\begin{figure*}[!h]
    \centering
    \includegraphics[width=.9\linewidth]{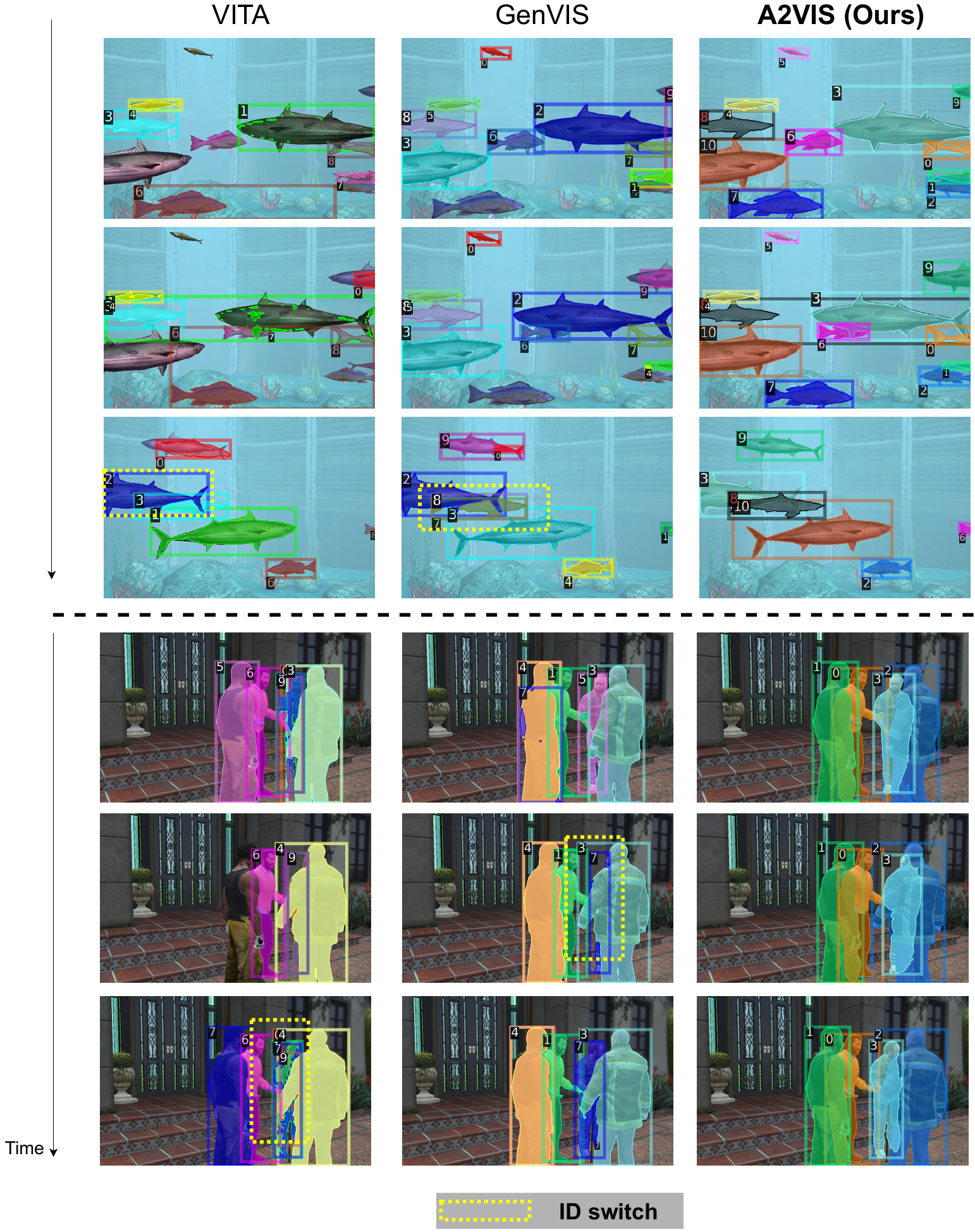}
    \caption{Qualitative comparison between our A2VIS and VITA and GenVIS on FISHBOWL dataset (top) and SAILVOS dataset (bottom). Instances with the same identity are consistently color-coded across all frames. }
    \label{fig:quali_compare_vis_all}
\end{figure*}

\begin{figure*}[!h]
    \centering
    \includegraphics[width=.9\linewidth]{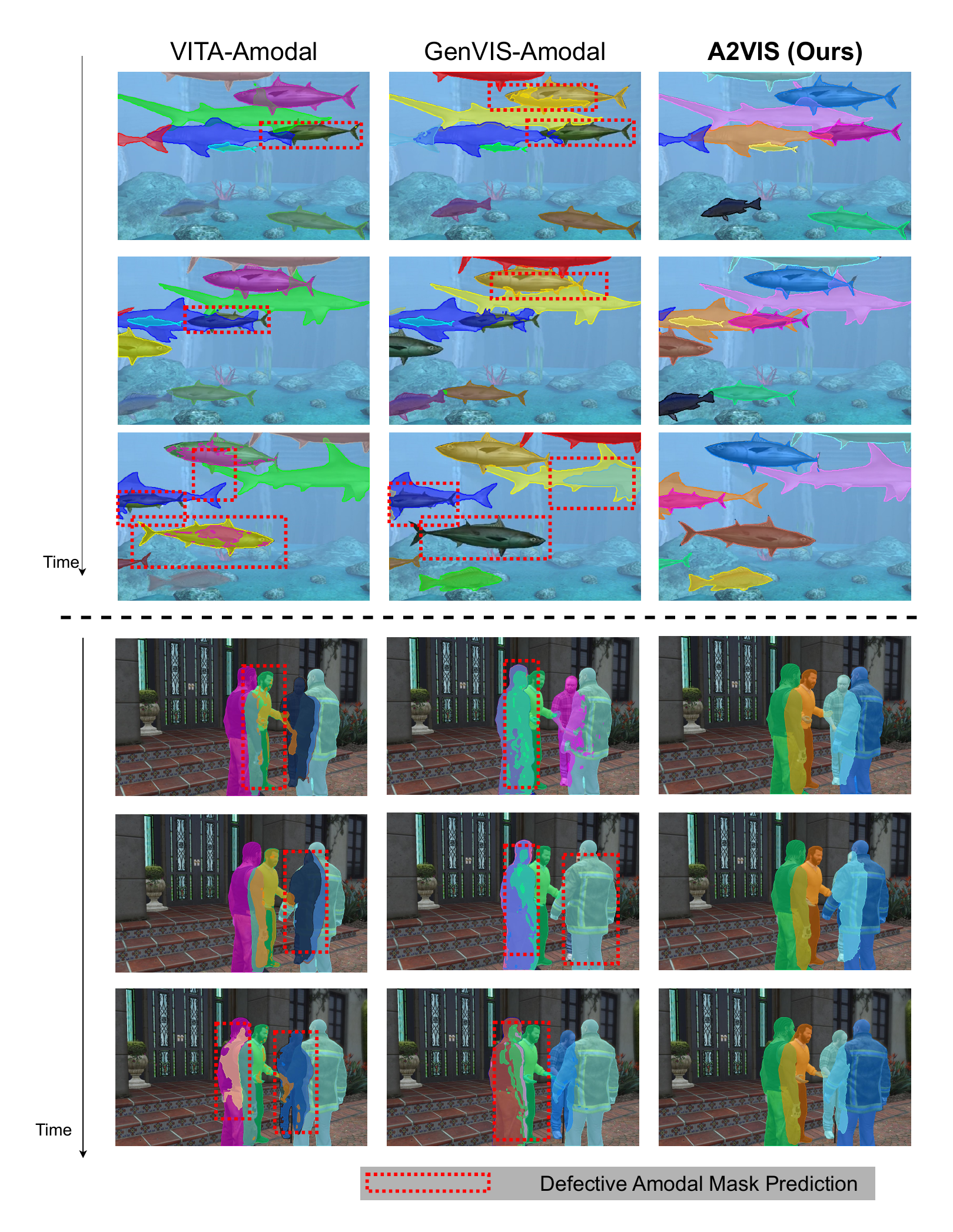}
    \caption{Qualitative comparison between our A2VIS and VITA-Amodal and GenVIS-Amodal on FISHBOWL dataset (top) and SAILVOS dataset (bottom). Instances with the same identity are consistently color-coded across all frames.}
    \label{fig:quali_compare_amodal_all}
\end{figure*}

\begin{figure}[!t]
    \centering
    \includegraphics[width=1\linewidth]{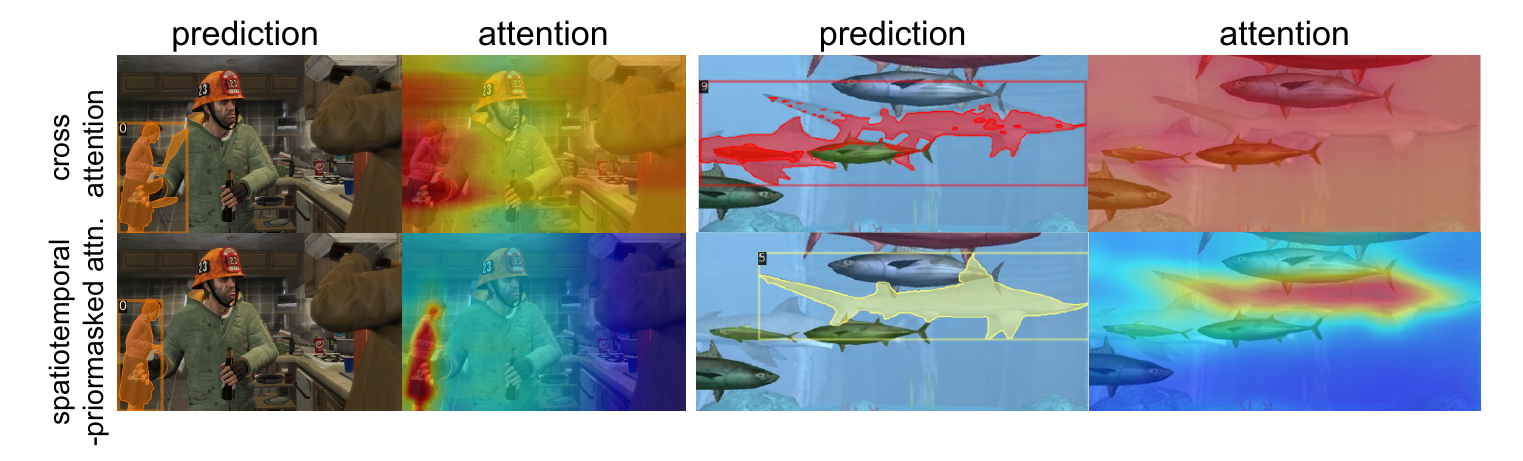}
    \caption{A visual comparison between using cross-attention and spatiotemporal-prior masked attention in SAMH.}
    \label{fig:attention_compare}
\end{figure}

\subsubsection{Comparison with {Amodal VIS} baselines.} 
Table \ref{tab:fishbowl_sailvos_amodal} compares A2VIS with baselines in amodal VIS. As depicted in the table, A2VIS consistently outperforms the baselines across various backbones, datasets, and metrics. 
Particularly, A2VIS achieves a significant performance advantage in segmentation tracking metric (AP and AR) over other amodal baselines
These results highlight the challenge of accurately predicting masks for occluded visual information in baseline methods. In contrast, A2VIS, with its proposed SAMH, clearly demonstrates its effectiveness in the task of amodal mask prediction.

\subsubsection {Comparison with MOT methods.} 

Table \ref{tab:track_compare} shows the comparison of A2VIS with MOT methods.
Here, \textit{TrackFormer} and \textit{MOTRv2} track objects using amodal bounding boxes. As can be seen, A2VIS achieves superior performance across metrics on both datasets. This suggests that A2VIS effectively mitigates the challenges associated with overlapping ambiguities in existing MOT methods. Moreover, it enhances the perception of complete instances even in the presence of occlusion, thereby facilitating the seamless and consistent tracking of objects.

\subsection{Qualitative Performance and Comparison.} 
Figure \ref{fig:additional_quali_results} illustrates qualitative performance of A2VIS on FISHBOWL dataset (top) and SAILVOS dataset (bottom). Moreover, further video qualitative results of A2VIS are provided as an .mp4 video in \href{https://drive.google.com/file/d/13AtQ9-RBN0kDkxQ0qCut59Y_k1UoiUEp/view?usp=sharing}{Link to video demo}

Figure \ref{fig:quali_result} visually compares between A2VIS and GenVIS-Amodal on the SAILVOS testset. A2VIS succesfully recognizes and maintains the identity of instances, even in scenarios where they are mostly occluded. 

Figures \ref{fig:quali_compare_vis_all} quatitatively illustrates the comparison between A2VIS and VIS baselines regarding tracking performance, namely VITA~\cite{heo2022vita} and GenVIS~\cite{heo2023generalized}  on FISHBOWL dataset (top) and SAILVOS dataset (bottom), respectively. As evident from these two figures, it is apparent that A2VIS, through the effective incorporation of amodal knowledge, operates at a superior level by acquiring the capability to perceive the complete trajectory and shape of a target. In contrast, VITA and GenVIS encounter a fundamental challenge at the level of occlusion, wherein they tend to perceive the previously tracked target as a new identity after being obscured. For example, in Figure \ref{fig:quali_compare_vis_all}, we highlight the ID switch cases of VITA and GenVIS in the dashed yellow boxes. This distinction places A2VIS on a different level compared to other VIS methods that lack the ability to predict occluded portions, rendering them prone to losing track of objects. 

Figure \ref{fig:quali_compare_amodal_all} depicts the qualitative comparison between A2VIS and the amodal VIS baselines, namely VITA-Amodal and GenVIS-Amodal. As shown in these two figures, A2VIS, coupled with the proposed SAMH, shows advantages compared with VITA-Amodal and GenVIS-Amodal in term of amodal segmentation.
The amodal segmentation results (e.g., fish, humans) produced by A2VIS is more consistent in comparison with VITA-Amodal and GenVIS-Amodal.
Moreover, we also observe that VITA-Amodal and GenVIS-Amodal frequently predict defective amodal masks. 
For examples, in Figure \ref{fig:quali_compare_amodal_all}, we highlight the defectiveness of amodal mask predictions from those baselines in dashed red boxes.

\subsection{Ablation Study}
\label{sec:ablation}

\begin{table}[!t]
\centering
\setlength{\tabcolsep}{6pt}
\caption{Ablation study of VSPM and ASPM in our SAMH regarding amodal VIS.}
% \vspace{-1em}
\label{tab:abla_masked_attn}
\resizebox{\linewidth}{!}{%
    \begin{tabular}{cc|cc|ccc|cc|ccc}
    \hline
        \multirow{2}{*}{\textbf{VSPM}} & \multirow{2}{*}{\textbf{ASPM}} & \multicolumn{5}{c|}{\textbf{FISHBOWL}} & \multicolumn{5}{c}{\textbf{SAILVOS}} \\ \cline{3-12}
        ~ & ~ & \textbf{AP}$\uparrow$ & \textbf{AR}$\uparrow$ & \textbf{HOTA}$\uparrow$ & \textbf{IDF1}$\uparrow$ & \textbf{IDs}$\downarrow$ & \textbf{AP}$\uparrow$ & \textbf{AR}$\uparrow$ & \textbf{HOTA}$\uparrow$ & \textbf{IDF1}$\uparrow$ & \textbf{IDs}$\downarrow$ \\ \midrule
        \xmark & \xmark & 35.97 & 26.57 & 44.22 & 54.66 & 3603 & 18.21 & 13.98 & 27.49 & 22.90 & 21012 \\ 
        \checkmark & \xmark & 38.24 & 27.12 & 46.65 & 54.69 & 3415 & 21.22 & 14.88 & 30.74 & 24.16 & 20345 \\ 
        \xmark & \checkmark & 36.96 & 26.87 & 45.78 & 53.74 & 3327 & 21.15 & 14.83 & 30.86 & 24.66 & 18349 \\ 
        \checkmark & \checkmark & \textbf{40.16} & \textbf{27.41} & \textbf{49.04} & \textbf{58.43} & \textbf{3275} & \textbf{23.41} & \textbf{15.04} & \textbf{32.12} & \textbf{26.41} & \textbf{16923} \\ \hline
    \end{tabular}
}
% \vspace{-1em}
\end{table}

\subsubsection{Impact of spatiotemporal-prior Masked Attention in SAMH.} 
Table \ref{tab:abla_masked_attn} evaluates the impact of Spatiotemporal-prior Masked Attention by considering SAMH with and without ASPM and VSPM, corresponding to the long-range and short-range prior knowledge, respectively. Incorporating either ASPM or VSPM leads to significant improvements in amodal VIS performance across all metrics. The combination of both ASPM and VSPM within SAMH achieves the best performance on both the FISHBOWL and SAILVOS datasets. This result validates the hypothesis on the spatiotemporal-prior knowledge for accurate amodal segmentation prediction.
In Figure \ref{fig:attention_compare}, we 
visualizes the attention map $\mathbf{T}^k + \mathbf{Q}\mathbf{K}^{\top} \in \mathbb{R}^{N_p\times N_c H_e W_e}$ (bottom row) in comparison with the traditional cross attention (top row). Among $N_p$ instance prototypes in the video, we only visualize the instance prototype that results in the segmentation mask highlighted. While traditional cross-attention spreads the attention map over the entire image, may overlooking the object of interest, spatiotemporal-prior Masked Attention module allows the model to focus on visible instance parts within a clip and global amodal segmentation, resulting in more precise and contextually relevant attention patterns.

\subsubsection{Length of clip ($N_c$) for training}
To determine the length of a clip in training, we performed $5$ runs and calculated their means. Table \ref{tab:abla_clip_len} shows the results of the ablation study on FISHBOWL and SAILVOS. In the online setting, where the clip length is set to $1$, A2VIS exhibited a decline in various scores, attributed to the absence of spatiotemporal-prior masked attention when $N_c = 1$, no reference frame is taken into account. Moreover, increasing the clip length to $5$ or $7$ did not necessarily improve performance.
Based on this empirical experiments, we selected a clip length of $3$ for training, as it yielded the highest scores. 
\begin{table}[!h]
\centering
\caption{Ablation study of the clip length ($N_c$) on FISHBOWL.}
\setlength{\tabcolsep}{9pt}
\renewcommand{\arraystretch}{1.1}
\label{tab:abla_clip_len}
\resizebox{.9\linewidth}{!}{%
    \begin{tabular}{c|cccc|cccc}
    \hline
        \multirow{2}{*}{\(N_c\)} & \multicolumn{4}{c|}{\textbf{Visible}} & \multicolumn{4}{c}{\textbf{Amodal}} \\ \cline{2-9}
        ~ & AP$\uparrow$ & AR$\uparrow$ & HOTA$\uparrow$ & IDF1$\uparrow$ & AP$\uparrow$ & AR$\uparrow$ & HOTA$\uparrow$ & IDF1$\uparrow$ \\ \midrule
        1 & 40.01 & 26.10 & 44.01 & 49.85 & 35.12 & 26.56 & 47.22 & 55.32 \\ 
        3 & 41.77 & 28.07 & 46.12 & 52.14 & 40.16 & 27.41 & 49.04 & 58.43 \\ 
        5 & 41.64 & 28.16 & 45.33 & 51.88 & 39.34 & 27.34 & 48.88 & 58.12 \\ 
        7 & 40.16 & 27.41 & 43.12 & 50.33 & 38.12 & 27.02 & 48.52 & 58.03 \\ \hline
    \end{tabular}
}
\end{table}

\subsubsection{Number of decoding layers $L$.}
We conducted an ablation study, as shown in Table \ref{tab:abla_num_L}, to assess the amodal VIS performance of the proposed SAMH across various decoding layer counts denoted by $L$. Similar to the earlier mentioned ablation study, we also provide the corresponding parameters required by SAMH for each $L$ value. The ablation study encompasses evaluations on FISHBOWL and SAILVOS datasets, reporting AP, AR and HOTA, IDF1. 
Our analysis concludes that a value of $L = 2$ strikes an optimal balance between performance and model complexity. Therefore, we have chosen to adopt $L = 2$ for A2VIS configuration.
\begin{table}[!h]
\centering
    \caption{Ablation study on the number of decoding layers $L$ in the proposed SAMH.}
    \label{tab:abla_num_L}
    \setlength{\tabcolsep}{9pt}
    \renewcommand{\arraystretch}{1.1}
    \resizebox{.9\linewidth}{!}{
    \centering
    \begin{tabular}{c|cccc|cccc}
    \toprule
        \multirow{2}{*}{$L$} & \multicolumn{4}{c|}{\textbf{FISHBOWL}} & \multicolumn{4}{c}{\textbf{SAIL-VOS}} \\ \cline{2-9}
        & AP$\uparrow$ & AP50$\uparrow$ & AP75$\uparrow$ & AR$\uparrow$ & AP$\uparrow$ & AP50$\uparrow$ & AP75$\uparrow$ & AR$\uparrow$ \\ \midrule
         1 & 37.23 & 58.34 & 39.22 & 27.11 & 21.03 & 28.75 & 17.21 & 14.72 \\ 
        2 & 40.16 & 59.60 & 42.35 & 27.41 & 23.41 & 31.11 & 19.12 & 15.04 \\
        3 & 40.34 & 59.23 & 42.44 & 27.40 & 23.22 & 31.23 & 18.31 & 15.04 \\ 
        5 & 40.92 & 60.12 & 42.33 & 27.43 & 23.56 & 31.44 & 19.11 & 15.06 \\ \bottomrule
    \end{tabular}
    }
\end{table}

\subsubsection{Number of convolutional layers of Amodal Feature Extraction}
In Section 3.5, we introduce the Amodal Feature Extraction $\Omega$, which extract the amodal mask feature $\mathbf{E}^k$ and the amodal attention feature $\mathbf{O}_k$. Here, $\Omega$ is designed by a sequence of convolutional layers ($3\times 3$ convolutional layers with a stride of $1$) where the first-half of the layers is responsible for outputting $\mathbf{O}_k$ whereas the second-half layers yields $\mathbf{E}^k$.
We empirically run with increasing number of convolutional layers complex as in Table \ref{tab:amodal_extractor_layer}). We thus choose 4 layers, which yields the best performance. 
\begin{table}[!h]
\centering
\setlength{\tabcolsep}{10pt}
\renewcommand{\arraystretch}{1.1}
    \centering
    \caption{Ablation study on the number of convolutional layers in Amodal Feature Extraction module.}
    \resizebox{0.46\linewidth}{!}{
    \begin{tabular}{c|ccc}
    \toprule
        \textbf{\#Layers} & \textbf{2} & \textbf{4} & \textbf{6} \\ \midrule
        AP & 40.12 & 41.77 & 41.01 \\ 
        AR & 27.67 & 28.07 & 26.98 \\ \bottomrule
    \end{tabular}}
    \label{tab:amodal_extractor_layer}
\end{table} 

\subsubsection{\revise{Impact of SAMH on VIS benchmark}}
\revise{Table \ref{tab:abla_w_and_wo_samh} presents the impact of SAMH on the VIS benchmark using the ResNet-50 model, comparing its performance on two datasets: FISHBOWL and SAILVOS. The results show that applying SAMH consistently improves performance across both datasets regarding VIS benchmark. For FISHBOWL, AP increases from 39.94 to 41.77, AR from 26.22 to 28.07, and IDF1 from 49.98 to 52.14, while the number of identity switches (IDs) decreases from 3493 to 3392, indicating better object tracking. Similarly, in SAILVOS, AP improves from 22.06 to 23.12, HOTA from 28.04 to 30.04, and IDF1 from 24.67 to 25.94, while the IDs decrease from 18142 to 17004, further demonstrating SAMH’s positive impact on tracking accuracy and stability. In general, these results highlight that SAMH improves both visible segmentation and tracking performance across different datasets.}
\begin{table}[!h]
    \centering
    \caption{\revise{Impact of SAMH on VIS benchmark using ResNet-50}}
    \setlength{\tabcolsep}{6pt}
    \renewcommand{\arraystretch}{1.1}
    \resizebox{.9\linewidth}{!}{
    \revise{\begin{tabular}{c|ccccc|ccccc}
    \toprule
        ~ & \multicolumn{5}{c|}{\textbf{FISHBOWL}} & \multicolumn{5}{c}{\textbf{SAILVOS}} \\ \midrule
        ~ & AP$\uparrow$ & AR$\uparrow$ & HOTA$\uparrow$ & IDF1$\uparrow$ & IDs$\downarrow$ & AP$\uparrow$ & AR$\uparrow$ & HOTA$\uparrow$ & IDF1$\uparrow$ & IDs$\downarrow$ \\ \midrule
        wo/ SAMH & 39.94 & 26.22 & 43.91 & 49.98 & 3493 & 22.06 & 15.27 & 28.04 & 24.67 & 18142 \\ 
        w/ SAMH & 41.77 & 28.07 & 46.12 & 52.14 & 3392 & 23.12 & 15.87 & 30.04 & 25.94 & 17004 \\ \bottomrule
    \end{tabular}}
    }
    \label{tab:abla_w_and_wo_samh}
\end{table}

\subsubsection{\revise{Impact of different occlusion levels}}
\revise{Table \ref{tab:abla_occ_rate} presents the performance of two methods, GenVis-Amodal and A2VIS, under two different occlusion rates: less than 50\% and greater than 50\%. For occlusion rates less than 50\%, A2VIS outperforms GenVis-Amodal across all metrics, with an AP of 42.33 compared to GenVis-Amodal's 38.62, and similarly higher values for AR, HOTA, and IDF1. A2VIS also has fewer identity switches (3123 vs. 3195 for GenVis-Amodal).}

\revise{In scenarios with occlusion rates greater than 50\%, A2VIS continues to show better performance than GenVis-Amodal, although the difference in scores is less pronounced. A2VIS achieves an AP of 33.14, while GenVis-Amodal reaches 29.78. Similar trends are observed for the other metrics, with A2VIS showing higher AR, HOTA, and IDF1 scores, and fewer identity switches. Overall, A2VIS consistently outperforms GenVis-Amodal, particularly under lower occlusion rates.}
\begin{table}[!h]
    \centering
    \caption{\revise{Impact of different occlusion levels on FISHBOWL, using ResNet-50}}
    \setlength{\tabcolsep}{12pt}
    \renewcommand{\arraystretch}{1.1}
    \resizebox{.9\linewidth}{!}{
    \revise{\begin{tabular}{l|l|ccccc}
    \toprule
        \shortstack{\textbf{Occlusion}\\ \textbf{Rate}} & \textbf{Method} & \textbf{AP}$\uparrow$ & \textbf{AR}$\uparrow$ & \textbf{HOTA}$\uparrow$ & \textbf{IDF1}$\uparrow$ & \textbf{IDs}$\downarrow$ \\ \midrule
        $<$50\% & GenVis-Amodal & 38.62 & 28.47 & 50.23 & 57.34 & 3195 \\ 
        ~ & A2VIS & 42.33 & 29.24 & 52.18 & 60.12 & 3123 \\ \midrule
        $>$50\% & GenVis-Amodal & 29.78 & 22.65 & 39.87 & 49.92 & 3756 \\ 
        ~ & A2VIS & 33.14 & 24.35 & 42.96 & 53.29 & 3657 \\ \bottomrule
    \end{tabular}
    }}
    \label{tab:abla_occ_rate}
\end{table}

\subsection{Video Amodal Segmentation Comparison}
In this section, we compare A2VIS with amodal video object segmentation methods (e.g. SaVos~\cite{yao2022self}, C2F~\cite{gao2023coarse}, EoRaS~\cite{fan2023rethinking}).
Note that these methods focus solely on single object amodal segmentation, using ground-truth visible segmentation across frames as input.
On the other hand, A2VIS is an end-to-end framework that simultaneously detect, track, visible segmentation, and amodal segmentation for multiple objects in videos. Table \ref{tab:compare_w_avos} shows the comparison regarding task-specific capabilities between A2VIS and existing amodal video object segmentation methods.

To ensure fairness, we utilize predicted visible segmentations from A2VIS on FISHBOWL as input for their trained amodal predictor.
Given that SaVos~\cite{yao2022self} is the only model with its trained model published on FISHBOWL, we solely compare A2VIS with SaVos, as shown in \revise{Table \ref{tab:amodal_segm_compare}.}
\begin{table*}[!h]
    \centering
    \setlength{\tabcolsep}{6pt}
    \renewcommand{\arraystretch}{1.1}
    \caption{Comparison on task-specific capabilities between existing amodal video object segmentation methods and A2VIS.}
    \resizebox{\linewidth}{!}{
    \begin{tabular}{l|c|c|ccc}
    \toprule
        \textbf{Methods} & \textbf{\shortstack{Additional \\Input Mask} }& \textbf{Object} & \textbf{Tracking} & \shortstack{\textbf{Visible}\\ \textbf{Segmentation}} & \shortstack{\textbf{Amodal}\\ \textbf{Segmentation}} \\ \midrule
        SaVos & \checkmark & Single object & \xmark & \xmark & \checkmark \\ 
        C2F & \checkmark & Single object & \xmark & \xmark & \checkmark \\ 
        EoRaS & \checkmark \ & Single object & \xmark & \xmark & \checkmark \\ \midrule 
        \textbf{A2VIS (Ours)} & \xmark & Multiple objects &  \checkmark & \checkmark & \checkmark \\ \bottomrule
    \end{tabular}}
    \label{tab:compare_w_avos}
\end{table*}

\begin{table}[!htb]
\centering
\setlength{\tabcolsep}{8pt}
\renewcommand{\arraystretch}{1.0}
    \centering
    \caption{Video amodal segmentation comparison on FISHBOWL}
    \label{tab:amodal_segm_compare}
    \resizebox{.6\linewidth}{!}{
    \begin{tabular}{l|llll}
    \toprule
        \textbf{Methods} & \textbf{AP}$\uparrow$ & \textbf{AP50}$\uparrow$ & \textbf{AP75}$\uparrow$ & \textbf{AR}$\uparrow$ \\ \midrule
        SaVos  & 38.21 & 55.61 & 41.32 & 27.33 \\ 
        \textbf{A2VIS (Our)} & 40.16 & 59.60 & 42.35 & 27.42 \\ \bottomrule
    \end{tabular}}
\end{table}

\subsection{Model complexity}
To benchmark computational complexity, we conducted inference on an RTX 8000 GPU using the Swin-L backbone across 20 test videos of FISHBOWL. A2VIS, with 222.8M parameters, averaged 0.77 FPS, which is competitive with GenVIS-Amodal's 220.3M parameters at 0.82 FPS. As in Table \ref{tab:complexity}, despite the competitive complexity, A2VIS shows significant gaps in performance in comparison with GenVIS-Amodal.
\begin{table}[!h]
\centering
\setlength{\tabcolsep}{8pt}
\renewcommand{\arraystretch}{1.0}
    \centering
    \caption{Model comparison between GenVIS-Amodal and our proposed A2VIS on tasks support, amodal performance and complexity.}
    \label{tab:complexity}
    \resizebox{\linewidth}{!}{
    \begin{tabular}{l|c|cc|cc}
    \toprule
        \multirow{2}{*}{\textbf{Methods}} & \multirow{2}{*}{\textbf{Tasks}} & \multicolumn{2}{c|}{\textbf{Performance}} &  \multicolumn{2}{c}{\textbf{Computational cost}}\\ \cmidrule{3-6}
        &  & \textbf{AP}$\uparrow$ & \textbf{IDS}$\downarrow$ & \textbf{Params}$\downarrow$ & \textbf{FPS}$\uparrow$ \\ \midrule
        GenVIS-Amodal & Amodal Segmentaiton & 40.66 & 3242 & 220.3M & 0.82 \\ 
        \textbf{A2VIS (Our)} &  Amodal Segmentaiton \& Visible Segmentaiton & 43.08 & 2547 & 222.8M & 0.77 \\ \bottomrule
    \end{tabular}}
\end{table}

\subsection{Evaluation on real-world dataset}
\begin{figure}
    \centering
    \includegraphics[width=1\linewidth]{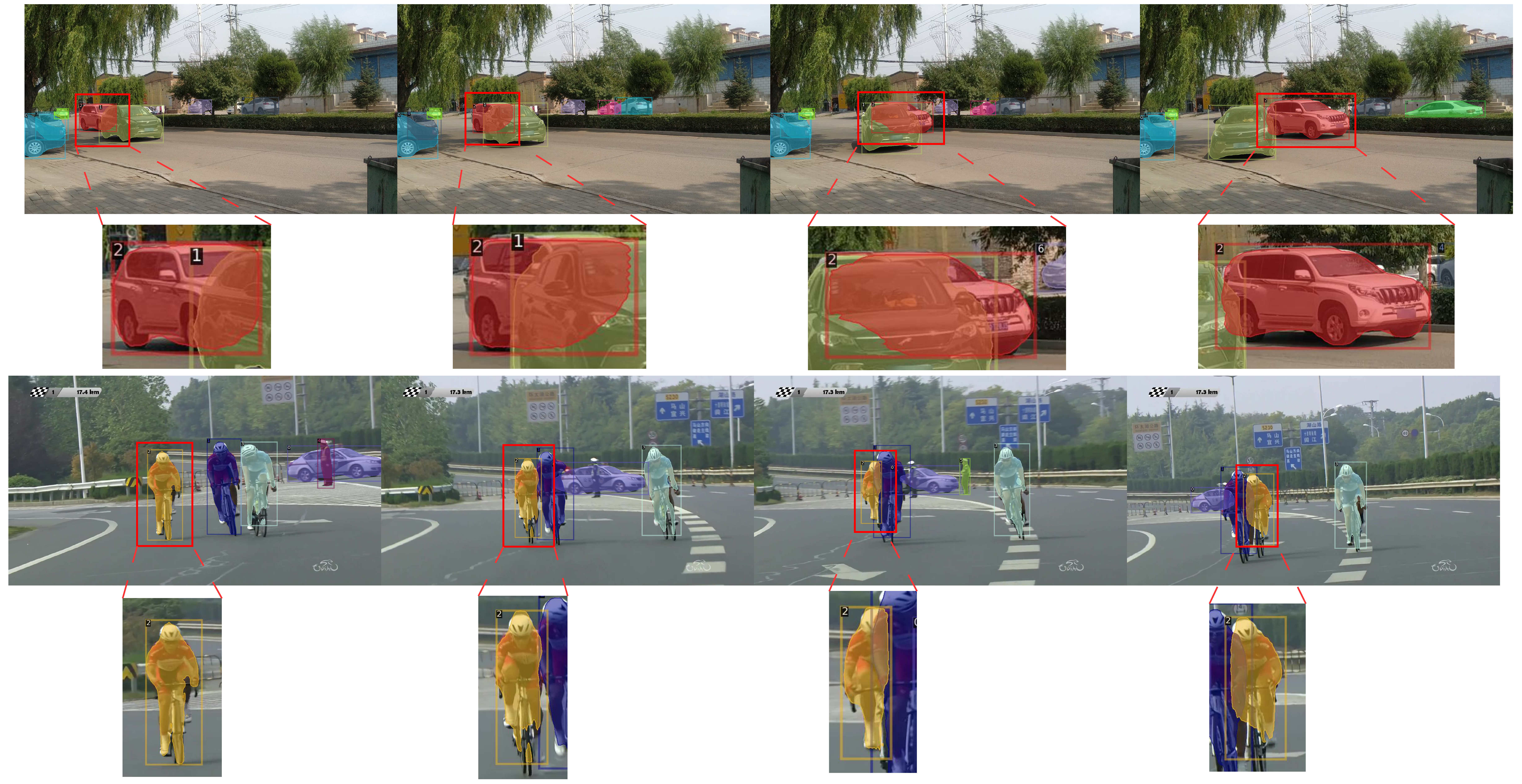}
    \caption{Qualitative outcomes of our A2VIS model, originally trained on SAILVOS, applied to inference on real-world dataset (videos are obtained from OVIS dataset). Best viewed in zoom and color.}
    \label{fig:ovis}
\end{figure}

Since there is currently no real-world Amodal VIS datasets available for comparison, we are unable to benchmark A2VIS on real-world scenarios at present.
We present a zero-shot evaluation on OVIS dataset for quantitative visible segmentation tracking, comparing our results with AISFormer-TrackRCNN, the only amodal VIS baseline predicting visible segmentation. Both AISFormer-TrackRCNN and A2VIS were trained on SAILVOS and then evaluated on OVIS regarding two categories (person \& car) without further training. Table \ref{tab:zeroshot_eval} shows that our A2VIS achieves significant improvement over the baseline. We also depict the qualitative results of A2VIS in Figure \ref{fig:ovis}. Although the amodal mask of the object may not be perfect, the inherent amodal awareness properties ensure the persistence of the object's presence even during occlusion, thereby preserving the identification of the object.

\begin{table}[!t]
    \centering
    \setlength{\tabcolsep}{8pt}
    \renewcommand{\arraystretch}{1.1}
    \caption{Zeroshot evaluation on OVIS dataset using Swin-L. }
    \resizebox{0.7\linewidth}{!}{
    \begin{tabular}{l|cccc}
    \toprule
        \textbf{Methods} & \textbf{AP}$\uparrow$ & \textbf{AP50}$\uparrow$ & \textbf{AP75}$\uparrow$ & \textbf{AR}$\uparrow$ \\ \midrule
        AISFormer-TrackRCNN & 9.01 & 17.33 & 7.02 & 8.00 \\ 
        \textbf{A2VIS} & 13.32 & 24.88 & 12.51 & 10.10 \\ \bottomrule
    \end{tabular}
    }
    \label{tab:zeroshot_eval}
\end{table}

\section{Conclusion and Discussion}
This paper introduces the innovative Amodal-Aware Video Instance Segmentation (A2VIS), a novel framework which utilize amodal characteristic into the processes of detection, segmentation, and tracking. A2VIS employs global instance prototypes to capture both visible and amodal characteristics of object in entire video, resulting in more robust object updates and association, especially in occluded scenarios. 
We also propose a Spatiotemporal-prior Amodal Mask Head (SAMH) for predicting amodal masks by utilizing both short-range and long-range spatiotemporal information.
Extensive experimentations and ablation studies conducted across benchmark datasets consistently highlight the superior performance of A2VIS compared to SOTA VIS methods
underscoring the significant benefits of A2VIS in the context of multiple object tracking. In summary, A2VIS represents a substantial advancement in video understanding, offering a versatile tool for tackling real-world scenarios involving object detection, segmentation, and tracking, especially in the presence of occlusion challenges. 

\noindent
\revise{\textbf{Limitation:}}
\revise{A2VIS attempts to reconstruct the occluded regions using visible cues from adjacent frames. Thus, objects undergoing large intrinsic shape changes are less suitable for A2VIS. Moreover, our method focuses on handling in-frame occlusions only. In particular, our approach does not explicitly account for objects that are occluded by being partly or completely out of the frame or disappear
in one frame and reappear in another. This limitation arises because existing amodal video instance segmentation datasets, such as FISHBOWL and SAILVOS, do not provide ground-truth annotations for objects that move out of the frame. As a result, we confine the amodal mask within the frame size.}

\noindent
\textbf{Discussion:} In future work, we aim to conduct studies on real-world datasets for amodal video instance segmentation. This will help to further validate and enhance A2VIS in practical scenarios as well as open new directions for research and exploration.

\noindent
\textbf{Acknowledgments} This material is based upon work supported by the National Science Foundation (NSF) under Award No OIA-1946391, NSF 2223793 EFRI BRAID and partly supported by Cobb Vantress Inc. 

\clearpage
\bibliographystyle{elsarticle-harv} 
\bibliography{main}

\end{document}